\relax

\documentclass[letterpaper]{article} 
\usepackage{aaai21}  
\usepackage{times}  
\usepackage{helvet} 
\usepackage{courier}  
\usepackage[hyphens]{url}  
\usepackage{graphicx} 
\usepackage{natbib}  
\usepackage{caption} 
\usepackage{amsmath}

\usepackage{booktabs}
\usepackage{times}
\usepackage{epsfig}
\usepackage{graphicx,subfigure}
\usepackage{amssymb}
\usepackage{bm}
\usepackage{makecell}
\usepackage{multirow}
\usepackage{wrapfig}
\usepackage{algorithm}
\usepackage{algorithmic}
\usepackage{adjustbox}
\usepackage{bbm}
\usepackage{epstopdf}
\usepackage{threeparttable}
\usepackage{tablefootnote}
\usepackage[normalem]{ulem}

\usepackage{color}
\usepackage[dvipsnames]{xcolor}

\DeclareMathAlphabet{\mathcal}{OMS}{cmsy}{m}{n}

\usepackage{amsmath}
\usepackage{amsthm}

\usepackage[switch]{lineno}

\newcommand{\name}{ColorCount}

\graphicspath{{figs/}}

\begin{document}

\title{Crowd Counting by\\
Self-supervised Transfer Colorization Learning
and Global Prior Classification}
\author {
    Haoyue Bai,
    Song Wen,
    S.-H. Gary Chan \\
}
\affiliations {
    Department of Computer Science and Engineering\\
    The Hong Kong University of Science and Technology, Hong Kong, China \\
    \{hbaiaa, swenaa, gchan\}@cse.ust.hk
}
\maketitle
\begin{abstract}
\begin{quote}
Labeled crowd scene images are expensive and scarce.
To significantly reduce the requirement of the labeled images,
we propose \name{}, a novel CNN-based approach by combining self-supervised transfer colorization learning and global prior classification to leverage the abundantly available unlabeled data.
The self-supervised colorization branch learns the semantics and surface texture of the image by using its color components as pseudo labels. The classification branch extracts global group priors by learning correlations among image clusters. Their fused resultant discriminative features (global priors, semantics and textures) provide ample priors for counting, hence significantly reducing the requirement of labeled images. We conduct extensive experiments on four challenging benchmarks. \name{} achieves much better performance as compared with other unsupervised approaches. Its performance is close to the supervised baseline with substantially less labeled data ($10\%$ of the original one).
\end{quote}
\end{abstract}

\section{Introduction}
Crowd counting is to estimate the number of closely packed objects in an image of unconstrained scene (for concreteness we use people as objects in this paper)~\cite{sindagi2018survey, zitouni2016advances, bai2020cnn}.  It has wide applications in public safety, people monitoring, and traffic management~\cite{onoro2016towards, lempitsky2010learning, chan2008privacy}. 
Despite much study, crowd counting remains a challenging problem due to severe occlusion, large scale variation, uneven distribution of people, etc.

Via density map regression, Convolutional Neural Networks (CNNs) based methods have recently been shown to be promising with multi-branch architecture, local global context fusion and attention mechanisms~\cite{zhang2016single, sam2017switching, cao2018scale, boominathan2016crowdnet, liu2018decidenet, kang2018crowd, sam2019almost, wang2019learning, liu2018leveraging, bai2019crowd, ma2020learning}.
However, the previous approaches are highly data-driven, i.e.,
they require 
voluminous amount of diverse labeled data in the training
process. These data are expensive due to intensive annotation.
Such labeling cost
is especially high for crowd images because each of the individual
targets has to be annotated.  This is the major reason that only a few
hundred annotated images are available in current crowd counting
datasets~\cite{idrees2013multi, zhang2016single, chan2008privacy}.
Such a small database is often not sufficient to achieve good transferability, leading to over-fitting and limiting their application
to diverse real-world scenarios.

While labeled crowd images are expensive and scarce, unlabeled ones are
widely available at virtually no cost.  In this work, we
study how to significantly reduce the need for labeled data by leveraging these abundant freely available unlabeled images as training data for crowd counting.
Our scheme, termed \name{}, is a novel approach based on self-supervised transfer colorization learning and global prior classification
to extract the discriminative features of crowds, so as to
markedly reduce overfitting and the need for costly labeled data.
Colorization is to hallucinate a plausible color version of a grayscale photograph~\cite{zhang2016colorful}.
The key idea is that semantics and local texture pattern as obtained in the coloration process of an image reflects the density of people in the region, and hence provides important clues to its people count. Though we do not know the exact count in the region, we may use the fact that the number of people in a tight texture region (typically high-density area) is likely to be higher than that in a loose texture region (typically low-density area) or background (non-people regions). Furthermore, we extract the global discriminative features for counting by conditioning on categorical counting group priors. This can be viewed as a coarse-to-fine process to further fine-tune the count.

\vspace{0.25in}
\begin{figure*}[t]
	\centering
	\subfigure[The semantics and local texture constrains.]{
		\includegraphics[width=0.485\textwidth, height=0.25\textheight]{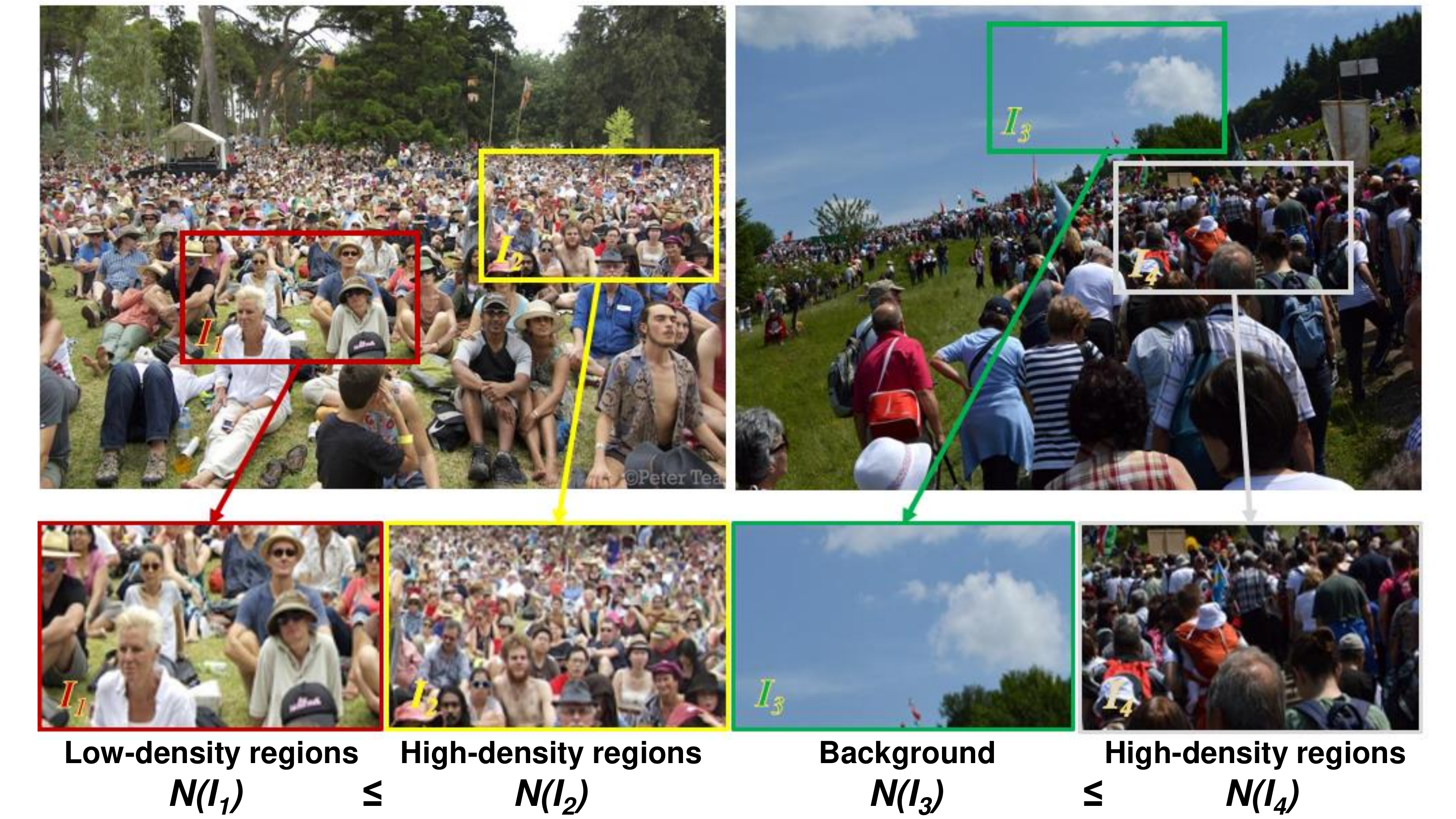}
		\label{fig:featureA}
	}	
	\subfigure[The global group priors.]{
		\includegraphics[width=0.485\textwidth, height=0.25\textheight]{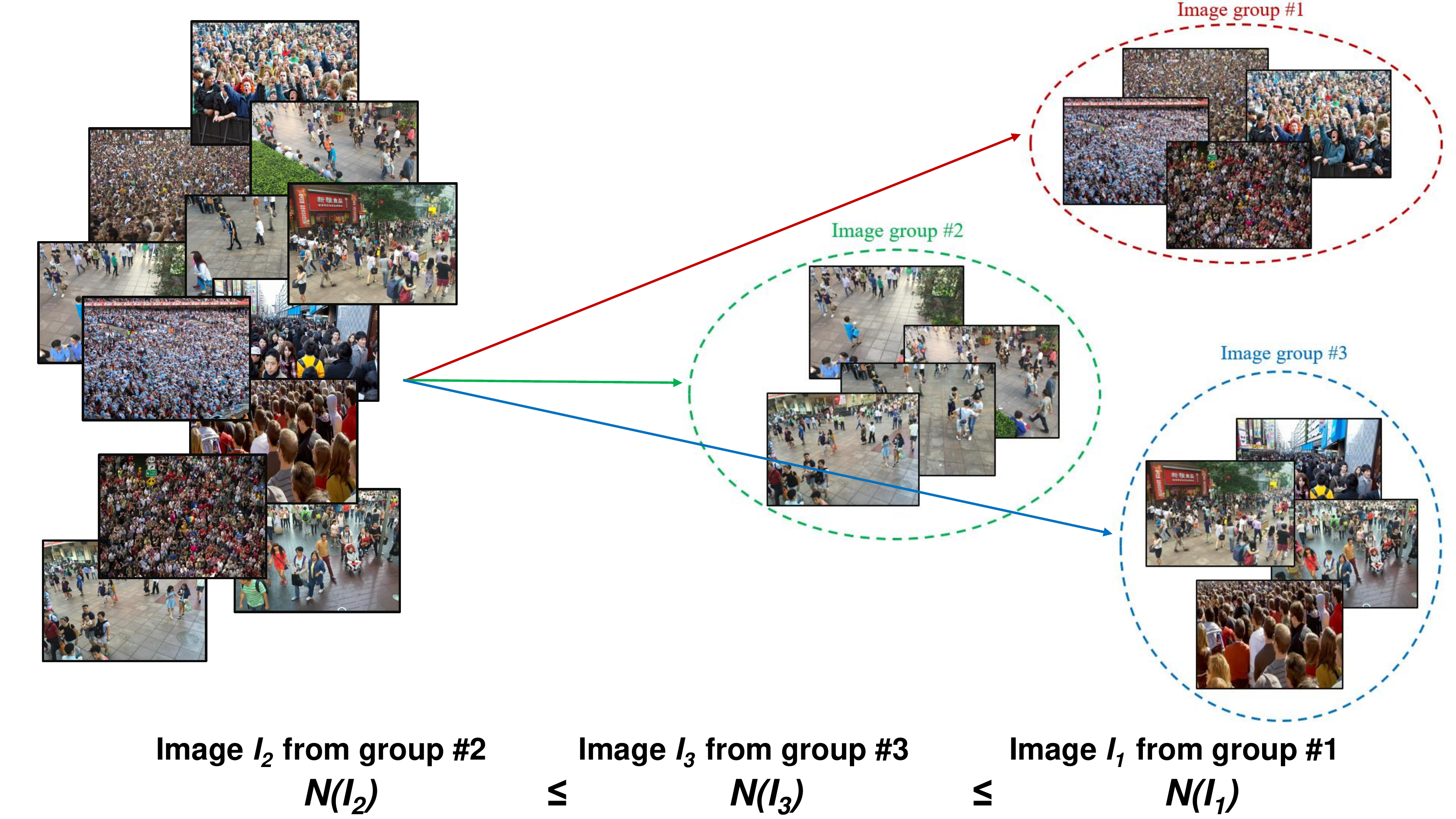}
		\label{fig:featureB}
	}
	\caption{Illustration of the discriminative features learned for counting: the local texture constrains (Figure a) and the global group priors (Figure b). $N(\cdot)$ denotes the number of people in a target crowd scene image. 
	}
	\label{fig:features}
\end{figure*}

\name{} consists of two sequential stages, self-supervised colorization pre-training using unlabeled data, followed by fine-tuning training using labeled data.
In pre-training, it has two branches, the colorization and classification branches, which fuse together to count the crowd. We illustrate the principles in Figure~\ref{fig:features}. 
The colorization branch exploits the process of colorization as an auxiliary task for counting and treating the color components of unlabeled images as the supervision signal (pseudo labels) to train its network. Self-supervised colorization via color loss and self-reconstruction extracts both the semantics and the surface texture of the scene in each unlabeled image. For example, the background sky is typically blue, and the background grass is typically green, etc.
As illustrated in Figure~\ref{fig:featureA}, the semantics and local textures provide clues on people count in the region. Let $N(\cdot)$ be the number of people in an image.
The number of people in tight region (high-density area) is larger than loose texture (low-density) region and background, i.e., $N(I_1) \le N(I_2)$ and $N(I_3) \le N(I_4)$.
The network learns the discriminative features of the image, which sheds insights on the count of the objects.

For the classification branch, though we do not know the exact number of people in each image in a group, the classification step would generally result in $N(I_2) \le N(I_3) \le N(I_1)$, where $I_j$ is any image in Group $\# j$ (Figure~\ref{fig:featureB}).
Noting that the colorization branch does not have global features of the image,
\name{} employs this classification branch which extracts
the global group priors by learning correlations among image clusters.
The two branches are combined with a joint loss to estimate the count. In other words, \name{} transfers the fused local and global knowledge learned from the colorization and classification process of the unlabeled images
to count the objects.
By leveraging the abundantly available unlabeled data, \name{} is much more scalable, flexible and applicable to general operating conditions.

To the best of our knowledge, this is the first piece of crowd counting work on joint self-supervised transfer colorization learning and global prior classification to leverage unlabeled images.
Using colorization as an auxiliary task and global prior classification,
\name{} jointly captures general semantics, counting-related local textual features and global
image group priors to learn discriminative features and achieve adaptivity on counting tasks.
We conduct extensive experiments on several public benchmark datasets and demonstrate that \name{} significantly reduces labeled datasets and achieves
much better performance given the same labeled dataset as compared with state-of-the-art unsupervised schemes.

This paper is organized as follows. We present related work
in Section~\ref{sect:related}, and
describe the details of \name{}
in Section~\ref{sect:colorcount}.
We conduct extensive experiments based on four real-world datasets and present the results in Section~\ref{sect:exp}.  We conclude in Section~\ref{sect:conclude}.

\section{Related Work}
\label{sect:related}

In this section, we review the crowd counting related literature. Section~\ref{rela:supervised} presents the supervised learning based approaches, and Section~\ref{rela:unsupervised} discusses utilizing unlabeled data for crowd counting.

\subsection{Supervised Learning for Crowd Counting} 
\label{rela:supervised}

Supervised learning based CNNs for corwd counting mainly focus on effective network design~\cite{zhang2016single, sam2017switching, cao2018scale, varior2019scale, idrees2013multi, boominathan2016crowdnet, liu2018decidenet, kang2018crowd, liu2018crowd, li2018csrnet, ma2021spatiotemporal, yang2020embedding}. MCNN proposes multi-column convolutional neural networks with different filter size to address the scale variation problem~\cite{zhang2016single}. Based on the MCNN, Switching-CNN designs a patch-based switching module with a multi-column structure which enlarges the scale range and better handles the scale variations~\cite{sam2017switching}. Besides, researchers also study stacking several multi-column blocks with densely upsampled layers to generate high-quality density maps~\cite{cao2018scale}.  

Crowdnet uses a combination of deep and shallow fully convolutional neural layers to predict the density map~\cite{boominathan2016crowdnet}. This can both extract high-level semantic information and low-level features effectively. AFP adopts an across-scale attention scheme to adaptive fuse different scales and adapts to scale changes within an image~\cite{kang2018crowd}. Researches have also incorporated LSTM modules into DRSAN to learn spatial information~\cite{liu2018crowd}. CSRNet proves that dilated convolutional operations can be used to enlarge the receptive field and promote accurate crowd estimation~\cite{li2018csrnet}. 

While impressive,
the approaches mentioned above require considerable diversified labeled data for training to reduce over-fitting. The labeling task for crowd images is especially expensive and tedious, since hundreds and even thousands of individuals are needed to be labeled in one image. Thus the current crowd counting datasets~\cite{zhang2016single, idrees2013multi, chan2008privacy, chen2012feature, idrees2018composition, zhang2016data, bai2021motion} are typically small, and cannot satisfy the needs of real-world applications. Our work is to substantially reduce the need for labeled data by leveraging freely available unlabeled crowd images.

\subsection{Utilizing Unlabeled Data for Crowd Counting}  
\label{rela:unsupervised}
Recently, leveraging unlabeled data in an unsupervised learning manner draws much attention. Some researchers have attempted to use unlabeled data for crowd counting. This is an alternative to address the over-fitting and reduce the demand for human annotations. GWTA-CCNN trains $99.9\%$ parameters of its model without any labeled data via reconstruction loss, and the remaining $0.1\%$ with supervision~\cite{sam2019almost}. However, the L2 reconstruction loss cannot fully extract discriminative features for counting tasks and will introduce much unrelated information during training. 
CCWId makes use of a large-scale synthesized dataset for crowd counting and uses CycleGAN to alleviate the domain gap~\cite{wang2019learning}. However, the distribution of synthesized people is different from real data. This results in a broader domain gap compared with using unlabeled real-world crowd scene images for counting.

Self-supervised learning is a subset of unsupervised learning methods. It learns visual features from large-scale annotation-free images via carefully designed auxiliary tasks~\cite{jing2019self}. L2R studies the use of ranking as a self-supervised pretext task for crowd counting~\cite{liu2018leveraging, liu2019exploiting}. However, ranking is a weak supervision signal for counting, which cannot fully extract discriminative features to learn to count. And this multi-task framework by minimizing an additional ranking loss is sensitive to the parameters. Carefully designed self-supervised pretext tasks can effectively learn useful visual features from large-scale unlabeled data for real-world crowd counting work. 

The labeling work for crowd counting task is expensive and tedious, while unlabeled images are cheap and abundant. Our target is to capture discriminative features for counting by utilizing unlabeled data and to reduce the need for intensive human annotation. Our proposed \name{} effectively extracts this information from unlabeled data by self-supervised colorization learning and global prior classification with local texture constraints and global categorical priors.

\begin{figure*}[!t]
	\centering
	\includegraphics[width=0.92\textwidth
	]{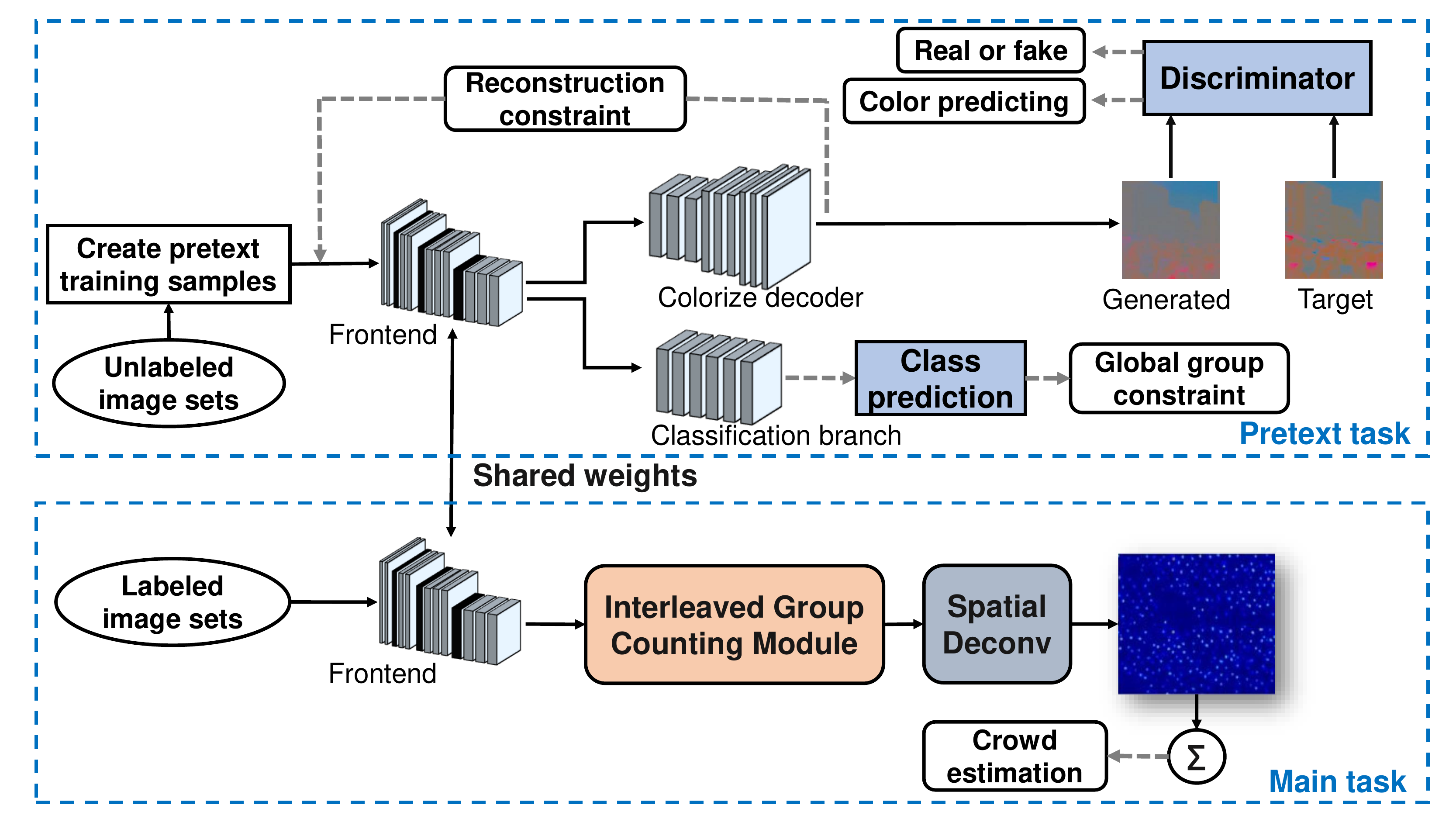}
	\caption{The framework of our proposed \name{}: Transfer Colorization Learning with Classification for Crowd Counting. The first row is the pre-training stage with pretext task. The second row is the fine-tuning stage for crowd counting (main task).
	}
	\label{fig:colorcount}
\end{figure*}

\section{\name{}: Transfer Colorization Learning with Classification for \\
Crowd Counting}
\label{sect:colorcount}

In this section, we describe the details of \name{}. In Section~\ref{meth:problem}, we present the problem formulation and baseline solution. Section~\ref{meth:color} shows how to use colorization as an auxiliary task given group priors for crowd counting.

\subsection{Problem Formulation and Baseline}
\label{meth:problem}

Labeled images for crowd counting are expensive because we need to label each of the individual targets within an image. In this paper, we propose \name{}, a novel self-supervised transfer colorization learning and global prior classification for crowd counting scheme to leverage unlabeled data and reduce the requirement for intensive human annotation. As Figure~\ref{fig:colorcount} shows, our \name{} contains two training stages: the pre-training stage with pretext task (the first row of Figure~\ref{fig:colorcount}) and fine-tuning stage with main task (the second row of Figure~\ref{fig:colorcount}).

In the first stage, we create pretext training samples from the unlabeled image sets. There are two branches in the pre-training stage: colorization branch and classification branch. The original unlabeled image is divided into a density map (as input image) and its color component (as the pseudo label). The network can be trained by taking L channel of the unlabeled image as input $X \in \mathbb{R}^{H \times W \times 1}$ and automatically generated color component as supervision signal $Y_{fake} \in \mathbb{R}^{H \times W \times 2}$ (ab color components in CIE Lab color space). Besides, we also pre-train the network with global group constraints. There are three ways to get group priors, as discussed in Section~\ref{sect:exp}.

In the second stage, we fine-tune the network with limited labeled data crowd counting. The network are fine-tuned with real counting label $Y_{real} \in \mathbb{R}^{H \times W \times 1}$. We propose an Interleaved Group Convolution based Crowd Counting Network (IGCCNet) as our baseline counting branch, which is consisted of three modules: the pre-trained frontend, the interleaved group counting module, and context fusion module. Besides, the counting branch is optimized with the Euclidean loss in our experiments, which is defined as: 
$\mathcal{L}_{E}=\frac{1}{N} \left ||F(X;\alpha)-Y_{real} \right ||_{2}^{2}$,
where $\alpha$ indicates the parameters, N is the number of pixels. The loss function in the fine-tuning stage, which is not limited to Euclidean loss, but can be any general function. We use Euclidean loss because it has been widely adopted in crowd counting with reportedly good performance, including our comparison schemes. For fairness, we hence use Euclidean loss here.

To be specific, the pre-trained frontend captures the transferred visual features. The interleaved group counting module is the main block to further enlarge the receptive field with limited parameters and effectively utilize counting features for density map prediction, which is a stack of interleaved group blocks~\cite{zhang2017interleaved}. The block sequentially contains two complementary interleaved group convolutions: primary group convolution with spatial convolution on each partition ($L$ primary partitions), and secondary group point-wise convolution ($M$ secondary partitions). The channels in each secondary partition are from different partitions in the primary group convolution. This operation is more efficient in terms of computation and parameters. Besides, the context fusion module fully utilizes the features extracted to achieve accurate crowd estimation.

\subsection{Training with Colorization Given \\Group Priors}
\label{meth:color}

We design \name{} to learn discriminative features for counting. Our \name{} is build based on the self-supervised colorization baselines. Besides, the texture constraint and categorical group constraint are included to effectively capture the local and global learning to count features. In this section, we present the details of our pre-training with colorization and given group priors process.

{\bf Colorization loss.} This is used for the first stage of \name{}. The original unlabeled image is divided into lightness channel and color component (as the pseudo label). The target is to learn a mapping function $\widehat{Y}= \mathcal{F}(X)$ from the lightness channel $X \in \mathbb{R}^{H \times W \times 1}$ to the other two ab color channels $Y \in \mathbb{R}^{H \times W \times 2}$ in the CIE Lab color space, where $\widehat{Y}$ indicates prediction, $Y$ denotes ground truth, and $H$, $W$ means image dimensions.

Instead of directly minimizing the Euclidean loss between $\widehat{Y}$ and $Y$, the value of both the lightness channel and the ab channels are quantized into grids. And then, we minimize a multinominal entropy loss between the two quantized color distributions to fully capture the semantic visual features~\cite{zhu2017unpaired},~\cite{zhang2017real}. For the input $X \in \mathbb{R}^{H \times W \times 1}$, we learn a mapping function $\widehat{Z}= \mathcal{G}(X)$ to a probability distribution of possible colors $\widehat{Z} \in [0, 1]^{H \times W \times Q}$. Then we minimize a multinomial entropy loss between predicted $\widehat{Z}$ ground truth color distribution $Z$ converted from ground truth color $Y$:
\small
\begin{equation}
\mathcal{L}_{cc}(\widehat{Z}, Z) = -\sum_{h,w}v(Z_{h, w})\sum_{q}Z_{h,w,q} \mathrm{log}(\widehat{Z}_{h,w,q}),
\end{equation}
\normalsize
where h, w is the image dimension, q indicates the number of quantized ab values, and $v(Z_{h,w})$ is a weighting term to rebalance the loss based on color-class rarity. Finally, the color value $\widehat{Y} \in \mathbb{R}^{H \times W \times 2}$ can be converted from the predicted probability distribution $\widehat{Z}$.

{\bf GAN loss.} To facilitate this pre-training process, we apply Cycle GAN~\cite{zhu2017unpaired} on two mapping functions $\mathcal{G} : X \rightarrow Z$ and $\mathcal{F} : Z \rightarrow X$. For the mapping function $\mathcal{G} : X \rightarrow Z$ and its discriminator $\mathcal{D}_Z$, the objective is

\small
\begin{equation}
\begin{aligned}
\mathcal{L}_{GAN}(\mathcal{G}, \mathcal{D}_Z, X, Z) = \mathbb{E}_{z \sim p_{data}(z)}[\mathrm{log}\mathcal{D_Z}(z)] +   \\
\mathbb{E}_{x \sim p_{data}(x)}[\mathrm{log}(1 - \mathcal{D_Z}(x)).
\end{aligned}
\end{equation}
\normalsize

\begin{figure*}[t]
	\centering
	\includegraphics[width=0.91\textwidth]{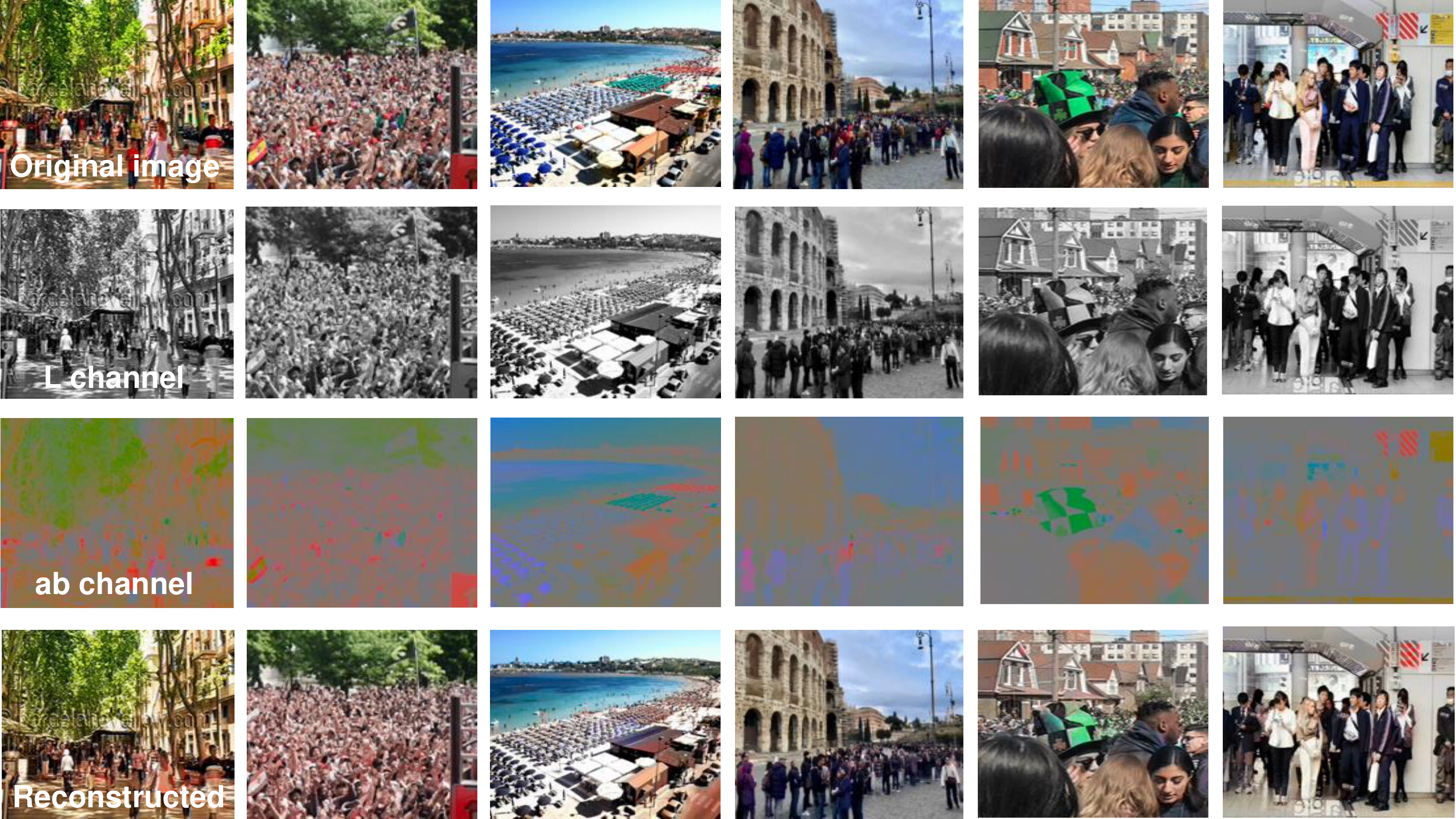}
	\caption{Visualization of learning from colorization based on our collected unlabeled data. The first line is the original image in the pretraining Dataset, the second line is the $L$ channel of the original image, the third line is the ground truth $ab$ channel, and the last line is our predicted colorization results.}
	\label{fig:colorvisual}
\end{figure*}

{\bf Self-recontruction loss.} The same to the mapping function $\mathcal{F} : Z \rightarrow X$ and its discriminator $\mathcal{D}_X$. This cycle reconstruction loss is:
\small
\begin{equation}
\begin{aligned}
\mathcal{L}_{cyc}(\mathcal{G}, \mathcal{F}) = \mathbb{E}_{x \sim   p_{data}(x)}[||\mathcal{F}(\mathcal{G}(x))-x||_{1}] +    \\
\mathbb{E}_{z \sim    p_{data}(z)}[||\mathcal{G}(\mathcal{F}(z))-z||_{1}].
\end{aligned}
\end{equation}
\normalsize

{\bf Texture loss.} To fully capture the local spatial information in the feature maps, texture loss term is adopted, which is widely used in style transfer~\cite{gatys2016image}. After mapped into the feature space based on a pretrained VGG-19 architecture, we compute the Gram matrices for both the output features. And then, we get the texture loss term, which is the mean squared error between the feature correlations of the computed Gram matrices. So, the texture loss is defined as:
\small
\begin{equation}
\mathcal{L}_{tex} = \frac{1}{4N^{2}_{l}M^{2}_{l}} \sum^{N_l}_{i=1}\sum^{N_l}_{j=1}(G^l_{i,j}-A^l_{i,j})^2,
\end{equation}  
\normalsize
where $N_l$ is the number of feature maps of VGG-19 layer $l$, $M_l$ is the product of layer $l$ feature maps height and width. The Gram matrix is the inner product of vectorized feature maps. The special features will be more significant, and the features with smaller element values will become smaller after the inner product. So, the effect of the Gram matrix is to magnify the characteristics of the data and to obtain the texture details.

{\bf Classification loss.} Given the global group priors, the images is categorized into several groups with different people density~\cite{haralick1973textural},~\cite{cirecsan2012multi}. This can be viewed as a coarse-to-fine process. Training the network with this classification loss enables the network to learn counting related discriminative features for the following fine-tuning crowd counting stage. The loss function is defined as:
$\mathcal{L}_{cla} = -\sum_i^n \sum_j^m u_{ij} \mathrm{log}(f(X_{ij}))$,
where $m$ is the number of groups, and $f(\cdot)$ is in the format of softmax probability.

Therefore, the full loss function is
\small
\begin{equation}
\begin{aligned}
\mathcal{L}(\mathcal{G}, \mathcal{F}, \mathcal{D}_X, \mathcal{D}_Z) = \mathcal{L}_{GAN}(\mathcal{G}, \mathcal{D}_Z, X, Z) +  \\ \mathcal{L}_{GAN}(\mathcal{F}, \mathcal{D}_X, Z, X) + \alpha \mathcal{L}_{cc}(\widehat{Z}, Z) + \\
\beta \mathcal{L}_{cyc}(\mathcal{G}, \mathcal{F})
+ \gamma \mathcal{L}_{tex} + \lambda \mathcal{L}_{cla}.
\end{aligned}
\end{equation}
\normalsize
After the pre-training stage, we fine-tune the network using limited labeled images via Euclidean loss in our experiment, as we have discussed in Section~\ref{meth:problem}.

\section{Experiments and Illustrative Results}
\label{sect:exp}

In this section, we discuss the experiment details and results to evaluate our approach. Section~\ref{expe:setting} presents implementation details and the datasets. In Section~\ref{expe:matrix}, we describe the evaluation metrics and our comparison schemes.
Section~\ref{expe:results} shows the illustrative results of our \name{} scheme. Section~\ref{expe:ablation} details the ablation study.

\begin{table*}[!h]
    \centering
	\caption{\small Statistics of the four labeled crowd counting datasets.}
	\label{table:dataset}
	\begin{adjustbox}{max width=0.89\textwidth}
		\begin{tabular}{lcccccc}
			\toprule
			\toprule
			Dataset &Average Resolution &Images Numbers &Total &Max &Min &Mean \\
			\midrule
			UCF-QNRF~\cite{idrees2018composition}        &2013 $\times$ 2902 &1,525 &1,251,642 &815 &49 &12,865 \\
			ShanghaiTech A~\cite{zhang2016single}  &589 $\times$ 868   &482   &241,677   &501 &33 &3,139  \\
			ShanghaiTech B~\cite{zhang2016single}  &768 $\times$ 1024  &716   &88,488    &123 &9  &578     \\
			UCF\_CC\_50~\cite{idrees2013multi}     &2101 $\times$ 2888 &50    &63,974    &1,279 &94 &4,543       \\
			\bottomrule
			\bottomrule
		\end{tabular}
	\end{adjustbox}
\end{table*}

\begin{table*}[!t]
    \centering
	\caption{\small MAE and MSE error on the ShanghaiTech B dataset with different amounts of training data and different initialization.}
	\label{table:supervised}
	\begin{adjustbox}{max width=0.89\textwidth}
		\begin{tabular}{lcccc}
			\toprule
			\toprule
			Method &Amounts of labeled data & Initialization &MAE &MSE \\
			\midrule
			MCNN~\cite{zhang2016single}          &$100\%$ &None   &26.4 &41.3 \\
			Switching-CNN~\cite{sam2017switching} &$100\%$ &None   &21.6 &33.4 \\
			ACSCP~\cite{shen2018crowd}         &$100\%$ &None   &17.2 &27.4\\
			DRSAN~\cite{liu2018crowd}         &$100\%$ &None   &11.1 &18.2 \\
			CSRNet~\cite{li2018csrnet}        &$100\%$ &ImageNet &10.6 &16.0\\
			\hline
			\textbf{Ours:} \name{} &Random sampling $10\%$  &Unlabeled data &14.33 &26.70\\
			\textbf{Ours:} \name{} &Random sampling $50\%$  &Unlabeled data &\textbf{8.77} &\textbf{14.12}\\
			\bottomrule
			\bottomrule
		\end{tabular}
		\end{adjustbox}
\end{table*}

\subsection{Implementation Details and Datasets}
\label{expe:setting}
The training stage one is optimized with a learning rate of $10^{-4}$, and the batch size is 25. In the second stage, we fine-tune our model with limited real-world labeled data. The weight of the first layer in the transferred channel-wise encoder is duplicated to three copies, which is to accommodate the three-channel input in the fine-tuning stage with labeled images. The optimization for this stage is Adam solver, with a $10^{-5}$ learning rate. Our framework is implemented with PyTorch 0.4.0, CUDA v9.0. The code will be released.

For the categorical group priors of the classification in the pre-training stage, there are three different approaches to generate the image group sets for learning to count. The following is the details of the three different methods:

\begin{itemize}
	\item 
{\em Ranking-based group priors:} cropping and sampling a decreasing sequence sub-images of the original images.
The number of people in the original images is larger than its sub-images, and this can be used as natural prior information for global group~\cite{liu2018leveraging},~\cite{wilcoxon1992individual}.
	\item 
{\em Clustering-based group priors:} using clustering-based methods to recognize and classify the image datasets into various groups based on the density features presented in the image~\cite{yang2010image}.
	\item 
{\em Classification-based group priors:} generating classification labels in the image keyword query stage. We can jointly use different degree adverbs as the keyword to build the datasets~\cite{wang2010locality}.    
\end{itemize}

There is a trade-off between the group annotation cost and the level of label noise. We can get the ranking-based group priors with no additional cost from the original unlabeled images, while the priors are weaker than the other two methods. The clustering-based algorithm is also a good way to get pre-clustering labels at low cost. Classification-based group priors generating categorical in the query process. This is practical in the real-word applications and easily obtained. Currently, we use this kind of classification-based group approach (the group number is set to $3$ in our experiments) and formulate it as a cross-entropy loss function, as shown in Section~\ref{meth:color}. Though there is a cost in low/med/high labeling, the cost is significantly lower than the labeling cost for original crowd counting tasks.

\begin{figure*}[t]
	\centering
	\includegraphics[width=0.92\textwidth]{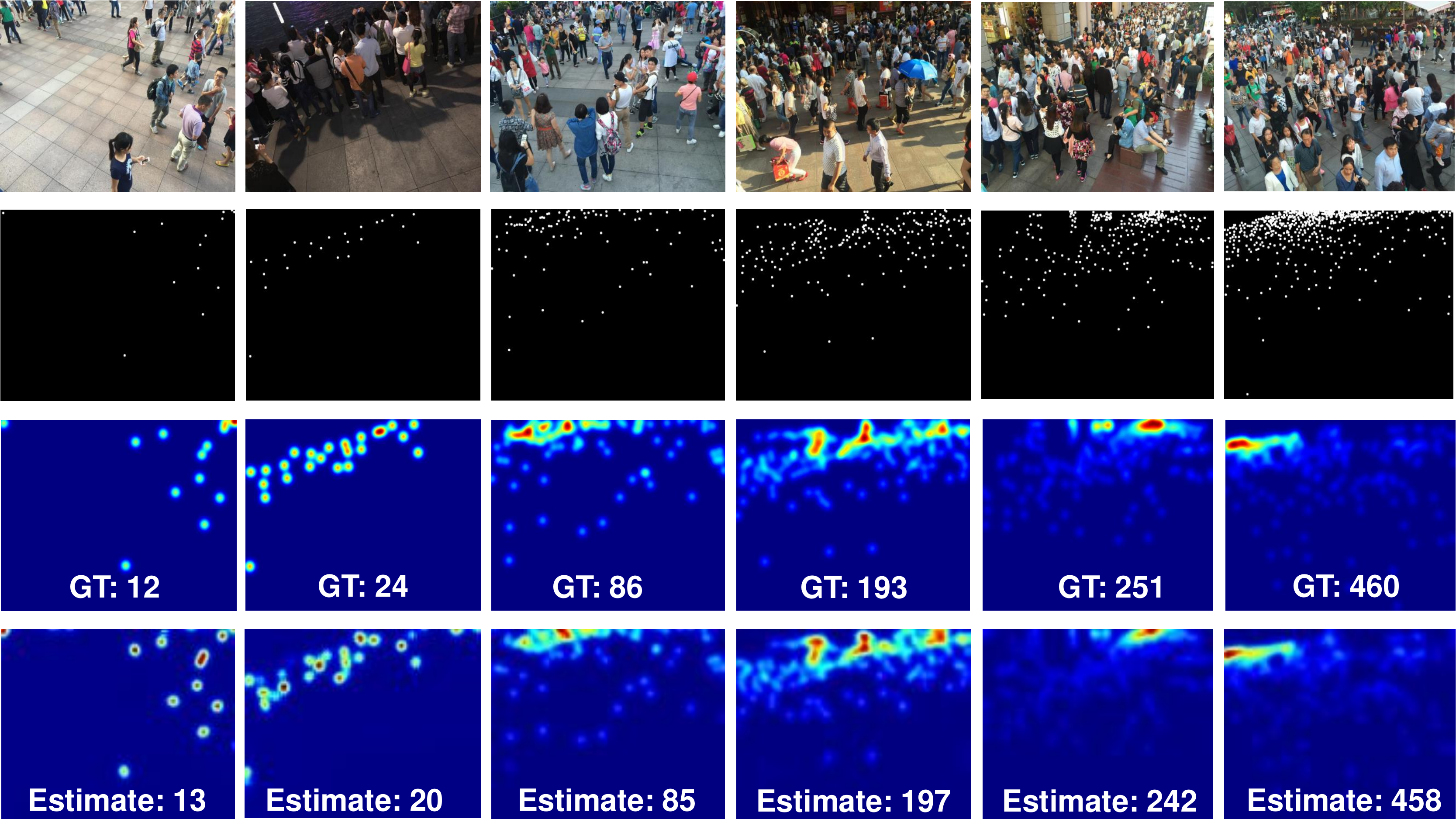}
	\caption{Qualititive results of the second training stage. The first row is original image, the second row is ground truth, and the third row is estimated density map.
	}
	\label{fig:countvisual}
\end{figure*}

\begin{table*}[!t]
    \centering
	\caption{\small Compared with other self-suervised learning based crowd counting methods by both leveraging unlabeled and labeled data.}
	\label{table:selfsupervised}
	\begin{adjustbox}{max width=0.96\textwidth}
		\begin{tabular}{lccccc}
			\toprule
			\toprule
			Method &Random sampling &Training process &Data collection &MAE &MSE \\
			\midrule
			L2R~\cite{liu2018leveraging}   &$60\%$ labeled data &Multi-task &Example query & 14.4 &23.8\\
			L2R~\cite{liu2018leveraging}   &$60\%$ labeled data &Multi-task &Keyword &13.7 &21.4\\
			\hline
			\textbf{Ours:} \name{} &$10\%$ labeled data &Pre-train $\&$ fine-tune &Keyword &14.33 &26.70\\
			\textbf{Ours:} \name{} &$50\%$ labeled data &Pre-train $\&$ fine-tune &Keyword &\textbf{8.77} &\textbf{14.12}\\
			\bottomrule
			\bottomrule
		\end{tabular}
		\end{adjustbox}
\end{table*}

For a fair comparison, We collect a large unlabeled dataset of crowd images from the internet by keyword query. This follows the same way of our comparison scheme L2R~\cite{liu2018leveraging, liu2019exploiting}, which is also making use of unlabeled data for crowd counting. We search from Google images with the keywords which have a higher probability of containing a crowd scene, such as Demonstration, Trainstation, Mall, Studio, Beach. Besides, we also jointly use different crowded degree adverbs at the keyword query to roughly classify the crowd scene images. We delete the images that not relevant to our problem. Finally, we collected a dataset with 2418 items of high-resolution crowd images. The total storage memory for this unlabeled dataset is 2.6GB.

We evaluate our \name{} on four challenging crowd counting datasets: UCF-QNRF~\cite{idrees2018composition}, ShanghaiTech A~\cite{zhang2016single}, ShanghaiTech B~\cite{zhang2016single}, and UCF\_CC\_50~\cite{idrees2013multi}. The statistics details of the four labeled crowd counting datasets are shown in Table~\ref{table:dataset}. To show the superior performance of our approach in reducing the requirement for labeled data, we randomly sample various levels of subsets (e.g., $10\%$, $50\%$) of all the four datasets for fine-tuning in our experiment. We conduct each experiment several times with different
random samples and the results show not much difference. Besides, we generate the ground truth for labeled crowd images via blurring each head annotation with a geometry-adaptive Gaussian kernel, as shown in Figure~\ref{fig:countvisual}.

\subsection{Evaluation Metrics and Comparison Schemes}
\label{expe:matrix}

For crowd counting, two metrics are used for evaluation, Mean Absolute Error (MAE) and Mean Squared Error (MSE), which are defined as:
\small
\begin{equation}
\begin{aligned}
\mathrm{MAE} = \frac{1}{N} \sum_{i=1}^{N}|C_i - \widehat{C}_i|,
\end{aligned}
\end{equation}
\begin{equation}
\begin{aligned}
\mathrm{MSE} = \sqrt{\frac{1}{N} \sum_{i=1}^{N}|C_i - \widehat{C}_i|^2},
\end{aligned}
\end{equation}
\normalsize
where $N$ is the total number of test images, $C_i$ means the ground truth count of the i-th image, and $\widehat{C}_i$ represents the estimated count.

The comparison schemes in our experiments are below:
\begin{itemize}
	\item 
{\em Supervised learning methods:} MCNN~\cite{zhang2016single}, Switching-CNN~\cite{sam2017switching}, ACSCP~\cite{shen2018crowd}, DRSAN~\cite{liu2018crowd}, CSRNet~\cite{li2018csrnet}, PGCNet~\cite{yan2019perspective}. These supervised learning based methods are trained on $100\%$ labeled data without unlabeled images.
\item 
{\em self-supervised learning methods:} L2R~\cite{liu2018leveraging}.  This method studies the use of ranking as a self-supervised pretext task for crowd counting and formulates it as a multi-task framework, which uses both labeled and unlabeled data. It uses two ways to collect unlabeled data: query by example and keyword.
\item 
{\em Almost unsupervised learning methods:} GWTA-CCNN~\cite{sam2019almost}. This scheme trains most of its parameters via unsupervised reconstruction loss, and only the last layers are trained with labeled data.
\end{itemize}

\subsection{Illustrative Results}
\label{expe:results}

In this section, we evaluate and analyze the results of our approach on four real-world labeled datasets. To show the superior performance of our approach to utilize unlabeled data and reduce the need for labeled images, we randomly sample various level subsets (e.g., $10\%$, $50\%$ in our experiment) of the original labeled data for training instead of directly adopting the original fully annotated dataset. It demonstrates the capability of our \name{} approach for real-world applications with limited labeled training data.

\begin{table*}[!t]
    \centering
	\caption{\small Ablation study on four challenging crowd counting datasets.}
	\label{table:ablation}
	\begin{adjustbox}{max width=0.96\textwidth}
	\begin{tabular}{lccccccccc}
	    \toprule
	    \toprule
		Method& Training process                                     &\multicolumn{2}{c}{UCF-QNRF}&\multicolumn{2}{c}{Part A}&\multicolumn{2}{c}{Part B}&\multicolumn{2}{c}{UCF\_CC\_50}\\
		\cline{3-10}
		~ &~ &MAE &MSE &MAE &MSE &MAE &MSE &MAE &MSE \\
		\midrule
		{\scriptsize GWTA-CCNN}\cite{sam2019almost} & Self-reconstruction &N/A  &N/A  &154.7 &229.4  &N/A &N/A &433.7 &583.3\\
		\hline
		IGCCNet &{\scriptsize Plus} \textbf{\name{}} ($10\%$) &244.2  &439.3  &73.6  &118.1  &14.3  &26.7  &316.0  &429.5       \\
		IGCCNet &{\scriptsize Plus} \textbf{\name{}} ($50\%$)  &\textbf{216.1} &\textbf{346.8}  &66.5  &\textbf{109.8}  &8.8  &14.1  &259.6  &375.4\\
		\hline
		MCNN~\cite{zhang2016single} & None ($100\%$)    &277.0  &N/A  &110.2  &173.2  &26.4  &41.3  &377.6  &509.1  \\
		MCNN~\cite{zhang2016single}   &{\scriptsize Plus} \textbf{\name{}}  ($100\%$)       &266.8  &460.6  &100.5  &156.3  &22.6  &32.5  &338.5  &461.2   \\
		CSRNet~\cite{li2018csrnet}  & ImageNet ($100\%$)    &N/A  &N/A  &68.2  &115.0  &10.6  &16.0  &266.1  &397.5     \\
	    CSRNet~\cite{li2018csrnet}  &{\scriptsize Plus} \textbf{\name{}}  ($100\%$)        &N/A  &N/A  &\textbf{65.9}  &112.3  &\textbf{7.4}  &\textbf{12.1}  &\textbf{236.3}  &\textbf{310.9}   \\
		\bottomrule
		\bottomrule
	\end{tabular}
	\end{adjustbox}
\end{table*}

{\bf Compared with supervised learning methods.} Table~\ref{table:supervised} presents the results of our \name{} compared with $6$ supervised learning approaches with different architectures and initialization methods. Our approach, using only $50\%$ of the labeled data, outperforms most of the listed supervised learning methods, which trains on $100\%$ labeled data. Even when only using $10\%$ labeled data, our \name{} yields MCNN~\cite{zhang2016single}, Switching-CNN~\cite{sam2017switching}, ACSCP~\cite{shen2018crowd} in terms of MAE and MSE. The good performance is in the cost of leveraging unlabeled data for our \name{}, while the unlabeled data is cheap and abundant. So, this clearly shows that our \name{} can largely reduce the demand for labeled crowd images by leveraging unlabeled data, and is more suitable for real-world applications.

{\bf Compared with self-supervised learning methods.} As shown in Table~\ref{table:selfsupervised}, we compare \name{} with other self-supervised learning based methods L2R~\cite{liu2018leveraging} by leveraging both labeled and unlabeled data. L2R uses $60\%$ labeled data in the experiment. For a fair comparison, we use even less amount of labeled data: $50\%$ and $10\%$ in our experiment. Besides, we follow a similar data collection way with L2R by utilizing keyword queries. Even with less labeled data utilization ($50\%$ labeled data, Keyword), our \name{} achieves much less the MAE and MSE value compared with the self-supervised L2R approach ($60\%$ labeled data, Keyword).

{\bf Compared with almost unsupervised learning method.} We compare our \name{} with the almost unsupervised learning methods GWTA-CCNN~\cite{sam2019almost} in the Table~\ref{table:ablation}. The value of the second row in Table~\ref{table:ablation} reports the results of GWTA-CCNN. We see that our approach significantly improves the crowd counting performance in terms of MAE and MSE. Our \name{} outperforms the almost unsupervised learning method by a large margin.

{\bf Qualitative results.} Figure~\ref{fig:colorvisual} shows the visualization of our unlabeled dataset and the color predicting results in the first training stage. The first row is the original image, the second row is $L$ channel of the original image, the third row is the ground truth $ab$ channel, and the last row is our generated color images. These qualitative results show that our \name{} achieves plausible colorization results, e.g., the sky is painted in blue, and the grass is colorized to green, which indicates that the network effectively captures semantic and visual texture features. It provides useful hints to the crowd counting task with limited labeled data. Figure~\ref{fig:countvisual} presents the qualitative results of the crowd counting task. The first row is the original image, the second row is ground truth, and the third row is generated density maps, which shows good counting results.

\subsection{Ablation Study}
\label{expe:ablation}

As Table~\ref{table:ablation} shows, we conduct an ablation study on four challenging labeled crowd counting datasets.  The first column list the baseline model we used in our experiment. The second column presents the pre-training process way in the first training stage and how much labeled images it uses in the second training stage.  Besides, the symbol $-$ in the table means that a certain experiment is not conducted in the related works, and we will not compare to it. IGCCNet is our designed counting branch in our \name{}. The experiments discussed above are mainly based on this network. The original MCNN and CSRNet is a fully supervised learning based crowd counting method, which only makes use of the labeled dataset. The original results are reported in the fifth row and seventh row.

Furthermore, we conduct experiments on MCNN plus \name{} and CSRNet plus \name{}, which means pre-trained on unlabeled data use our designed pre-training scheme and then fine-tune with $100\%$ labeled images. We show that the results of plus \name{} is always better than the results with only labeled data. Finally, we show that our approach can achieve superior performance with limited labeled data, and our self-supervised transfer colorization learning method can largely reduce the requirements for crowd counting data annotation and can promote the crowd counting task to real-world applications. 

The unlabeled data can be accessed with virtually no cost, in contrast with the expensive annotation process of crowd counting tasks. Therefore, our focus is on reducing the requirement for labeled crowd counting images. Table~\ref{table:supervised} and Table~\ref{table:ablation} show that \name{} achieve the same results with much less labeled data. Noted that it is difficult to compare with the fully supervised baseline methods. Therefore, we design two sets of experiments for a fair comparison with supervised baselines: 1) ColorCount with much less labeled data (Table~\ref{table:supervised}, Table~\ref{table:ablation}); and 2) ColorCount with the same amount of labeled data (Table~\ref{table:ablation}).

\section{Conclusion}
\label{sect:conclude}
In this paper, we propose \name{}, a novel self-supervised transfer colorization learning with group prior classification approach to leverage widely available annotation-free images for the crowd counting task. \name{} then uses colorization as a proxy task to learn discriminative features (general semantic information, local textual features) for learning to count and jointly extract global image group priors using classification.
The fused both global priors and local texture information provide important and ample knowledge as discriminative counting features with excellent transferability. As a result, the second training stage is able to fine-tuned network parameters with much reduced labeled crowd counting data. Extensive experiments on four challenging datasets show our superior performance. Our proposed \name{} achieves much better results in terms of MAE and MSE compared with other unsupervised learning methods and is close to the supervised baselines with much less labeled data.

\clearpage

\bibliography{main}

\clearpage
\onecolumn
\appendix

\begin{figure*}[!t]
	\centering
	\includegraphics[width=0.99\textwidth]{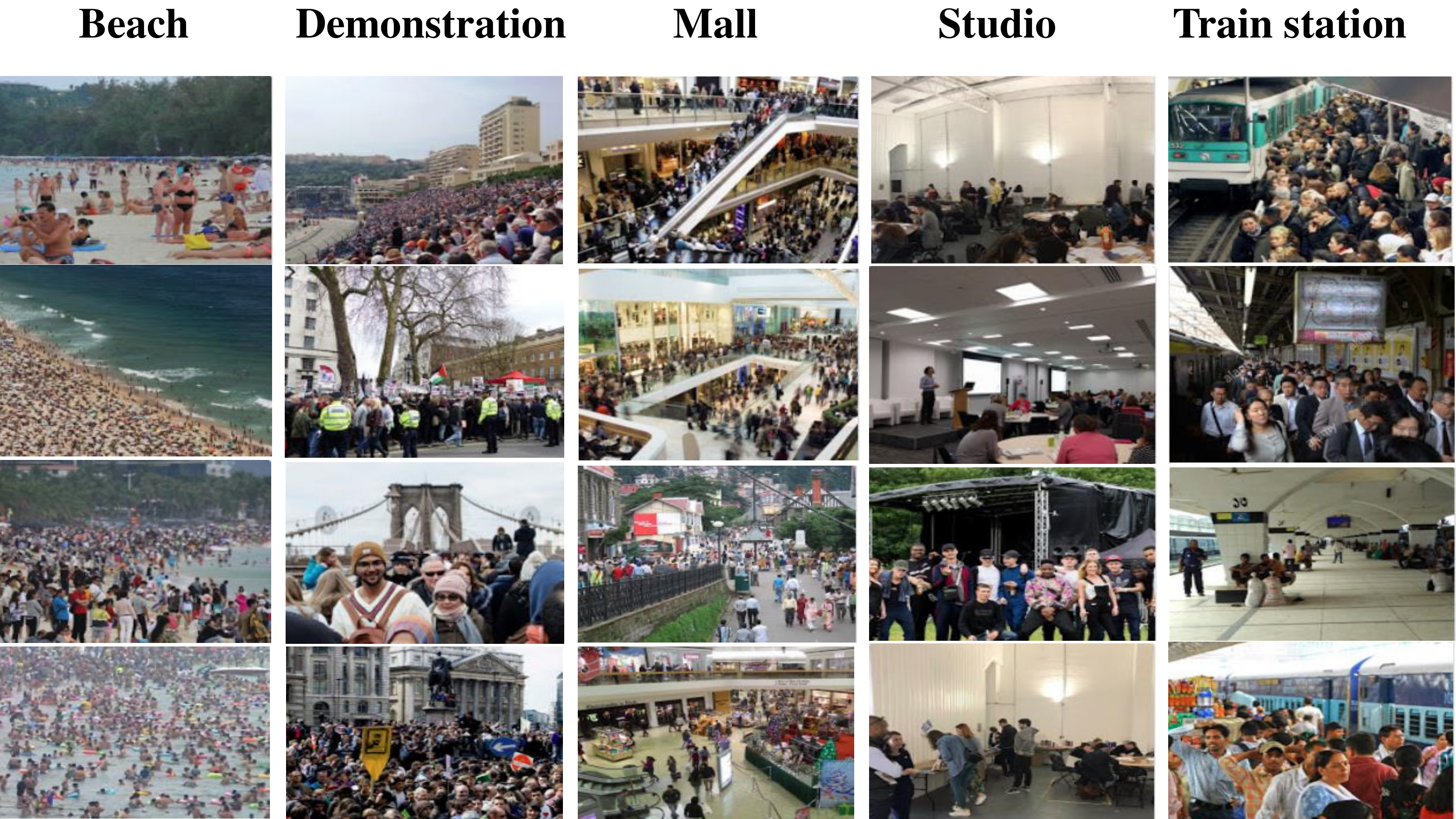}
	\caption{Unlabeled crowd counting dataset for pretraining collected by Keyword query. The first column is the example crowd images by keyword beach, the second column presents the images from quering the keyword demonstration, the third column shows the crowd example in mall environment, the fourth column is the example crowd images from keyword studio, and the last column shows the crowd images from train station.
	}
\end{figure*}

\begin{figure*}
	\centering
	\includegraphics[width=0.99\textwidth]{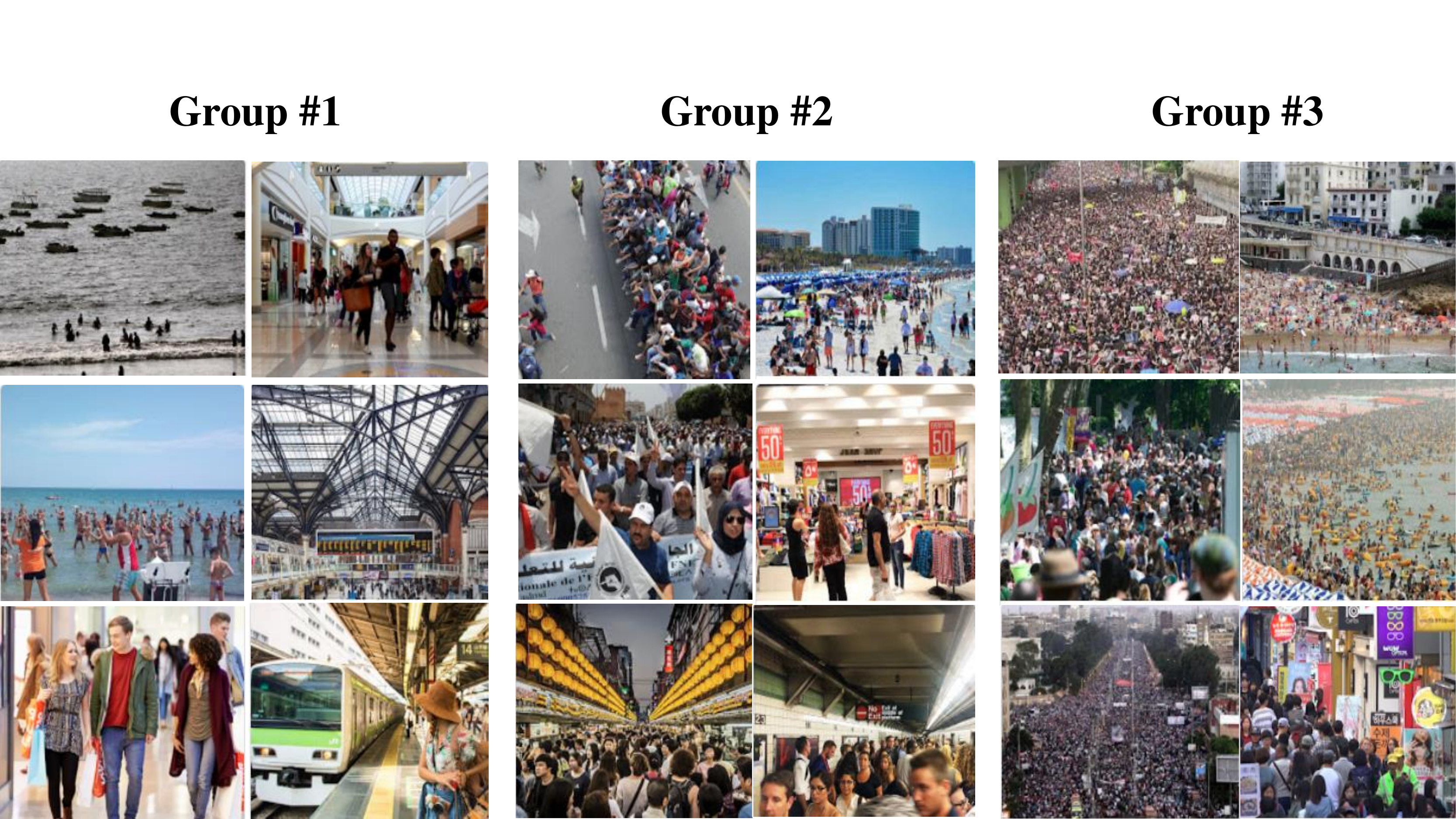}
	\caption{Group priors example images. The number of groups is three in our example. The first two columns indicate the images for group $\# 1$, which is the low crowd group. The third column and the fourth column shows the images for gourp $\# 2$, which means the middle crowd group. The last two columns presents the crowd images for group $\# 3$. And this group is the highly crowded group with relatively largest average people number for each scene.
	}
\end{figure*}

\begin{figure*}
	\centering
	\includegraphics[width=0.99\textwidth]{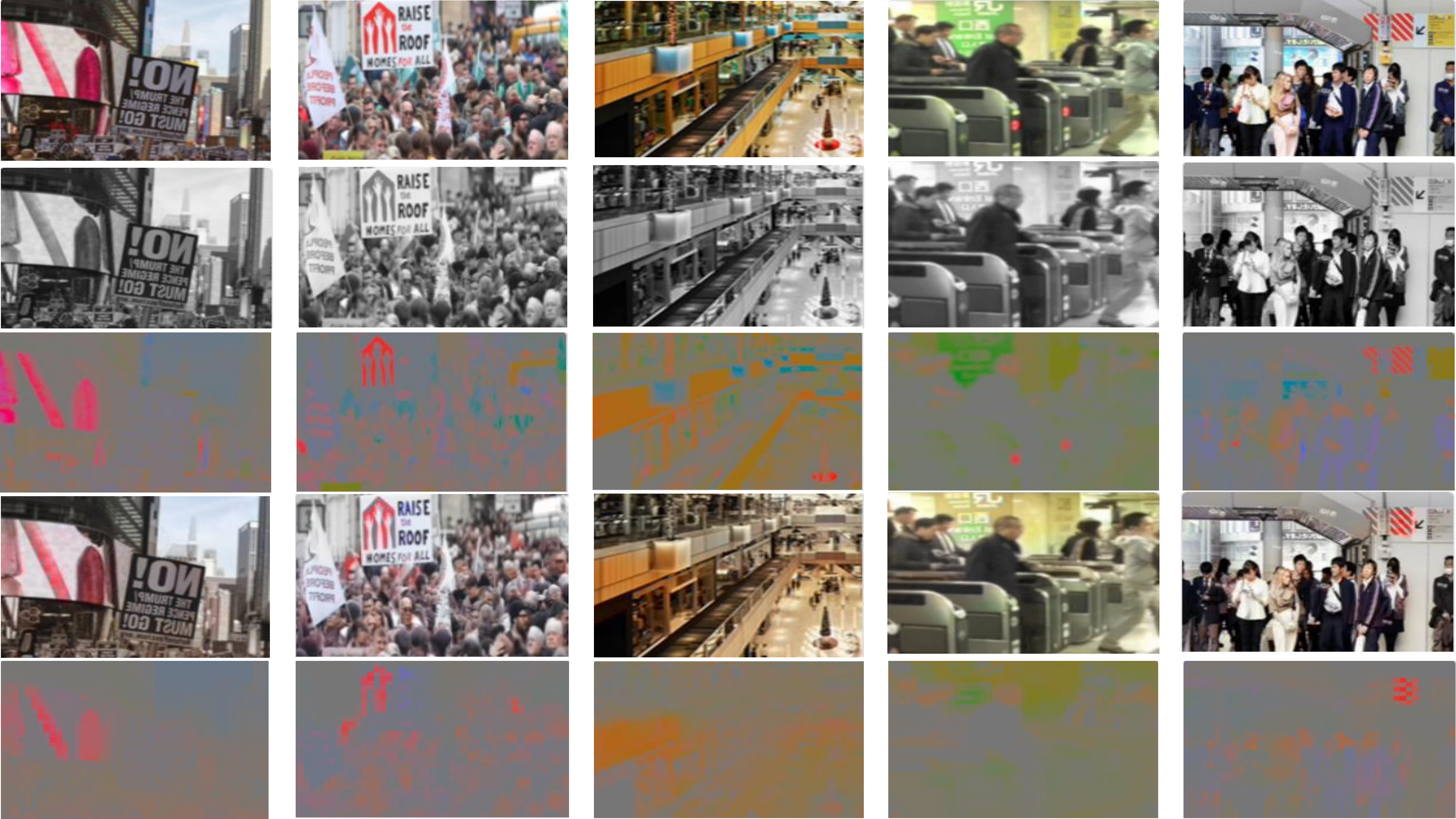}
	\caption{More examples for the visualization of colorization results on the unlabeled crowd images. The first row is the original unlabeled crowd images from keyword query, the second row shows the intensity map of the original crowd image, the third row is its corresponding color component, the fourth row presents the final color predicting result, and the last row is the predicted color component.
	}
\end{figure*}

\end{document}